# Spatio-Temporal Encoding and Decoding-Based Method for Future Human Activity Skeleton Synthesis

Tingyu Liu[1,*]  Jun Huang[2]  Chenyi Weng[1]
1. School of Mechanical Engineering, Southeast University, Nanjing 210096, China;
2. School of Mechanical Engineering, Nanjing University of Science and Technology, Nanjing 210094, China

**Abstract:** Inferring future activity information based on observed activity data is a crucial step to improve the accuracy of early activity prediction. Traditional methods based on generative adversarial networks(GAN) or joint learning frameworks can achieve good prediction accuracy under low observation ratios, but they usually have high computational costs. In view of this, this paper proposes a spatio-temporal encoding and decoding-based method for future human activity skeleton synthesis. **Firstly**, algorithms such as time control, discrete cosine transform, and low-pass filtering are used to cut or pad the skeleton sequences. **Secondly**, the encoder and decoder are responsible for extracting intermediate semantic encoding from observed skeleton sequences and inferring future sequences from the intermediate semantic encoding, respectively. **Finally**, joint displacement error, velocity error, and acceleration error, three higher-order kinematic features, are used as key components of the loss function to optimize model parameters. **Experimental results** show that the proposed future skeleton synthesis algorithm performs better than some existing algorithms. It generates skeleton sequences with smaller errors and fewer model parameters, effectively providing future information for early activity prediction.

## 0  Introduction

Early Activity Prediction[1, 2] Algorithms aim to infer and classify activities based on partially observed activity segments before the activity is completed. The absence of future information in partially observed activities makes early activity prediction more challenging than traditional human activity recognition tasks.

In related works, researchers have attempted to use algorithms such as one-shot mapping[3] and transfer learning[4-6] to achieve early activity prediction, which do not directly generate future activity information. However, these methods often struggle to ensure sufficient prediction accuracy and efficiency[7]. Therefore, inferring future activity information based on observed activity data is a crucial step to ensure the efficiency and accuracy of early activity prediction tasks.

Domestic and international scholars have also explored the generation of future activity information. Li et al.[8] proposed a Future Feature Supplement Network, where the features of future segments are used as labels. Through Generative Adversarial Network(GAN), they mapped the features of partially observed segments to future features, which were then fused with initial features and directly input into the classifier. Gammulle et al.[9] proposed a joint learning framework that simultaneously performs two tasks: generating future visual and temporal representations and early activity prediction. This joint learning framework ensures that the generated future representations positively guide the early activity prediction task. Chen et al.[10] proposed a cyclic semantic-preserving generative method for activity prediction, which learns to capture trends in partially observed sequences and uses GAN under certain constraints to supplement subsequent actions, thereby maintaining semantic consistency between the generated sequence and the observed sequence.

Although the aforementioned methods, whether based on GAN or joint learning frameworks, achieve high prediction accuracy under relatively low observation ratios, they often incur high computational costs. Therefore, constructing a lightweight network for generating future activity information is an important research of this paper.

With the rapid development of deep vision sensing technology[11], the skeleton sequence modality has become the main data carrier for human activity, which contains rich 3D information of the human body. Skeleton sequences encode the motion trajectories of human joints and typically have advantages such as small data volume, high computational efficiency, invariance to camera viewpoints, and robustness to clothing texture and background changes. Therefore, this paper proposes to use skeleton sequences as the data carrier for human activities.

This paper proposes a lightweight encoder-decoder framework-based future skeleton generation network (Future Skeleton Generation Network, FSGN) based on spatio-temporal encoding and decoding techniques. It comprises three parts: an input/output control module, an encoder-decoder network structure, and a network training loss function. Firstly, algorithms such as time control, discrete cosine transform, and low-pass filtering are used to cut or pad skeleton sequences. Secondly, the encoder and decoder are responsible for extracting intermediate semantic encoding from observed skeleton sequences and inferring future sequences from the intermediate semantic encoding, respectively. Finally, joint displacement error, velocity error, and acceleration error, as three higher-order kinematic features, are used as key components of the loss function to optimize model parameters.

## 1 Spatio-Temporal Encoding and Decoding-Based Method for Future Human Activity Skeleton Synthesis

To address the issue of high computational complexity in existing methods for synthesizing future skeleton sequences[12-16], this paper proposes a lightweight spatio-temporal encoding and decoding-based future skeleton generation network architecture, as shown in Figure 1.

The FSGN network architecture consists of three main components: an input/output control module, an encoder-decoder network structure, and a network training loss function:

**(1)Input Control Module:** This module consists of three parts: time control, discrete cosine transform, and low-pass filtering. Time control cuts or pads the historical skeleton sequence according to a set threshold to ensure that the sequence input into the encoder has a consistent length. The discrete cosine transform(DCT)[14] converts time-domain or spatial-domain signals to the frequency domain, achieving compression of skeleton sequence data. The low-pass filter (LPF) specifies a cutoff frequency to block signals above this frequency. The output control module consists of inverse discrete cosine transform and time control. The inverse discrete cosine transform converts the activity data back to the original domain, while time control ensures the length consistency of observed and future activity data.

**(2)Encoder and Decoder:** These are symmetrical structures, with the smallest unit being an improved MLP[17]. Compared to traditional MLP, the improved MLP replaces the activation function layer with an LN layer and adds residual connections, making it more suitable for Seq2Seq tasks.

**(3)Loss Function:** The loss function is defined as the sum of joint displacement error, velocity error, and acceleration error. By using higher-order kinematic features, it effectively constrains the spatio-temporal relationships during the generation of future activity information and optimizes the model parameters.

To address the issue of high computational complexity in existing methods for synthesizing future skeleton sequences, this paper proposes a lightweight spatio-temporal encoding and decoding-based future skeleton generation network architecture, as shown in Figure 1.

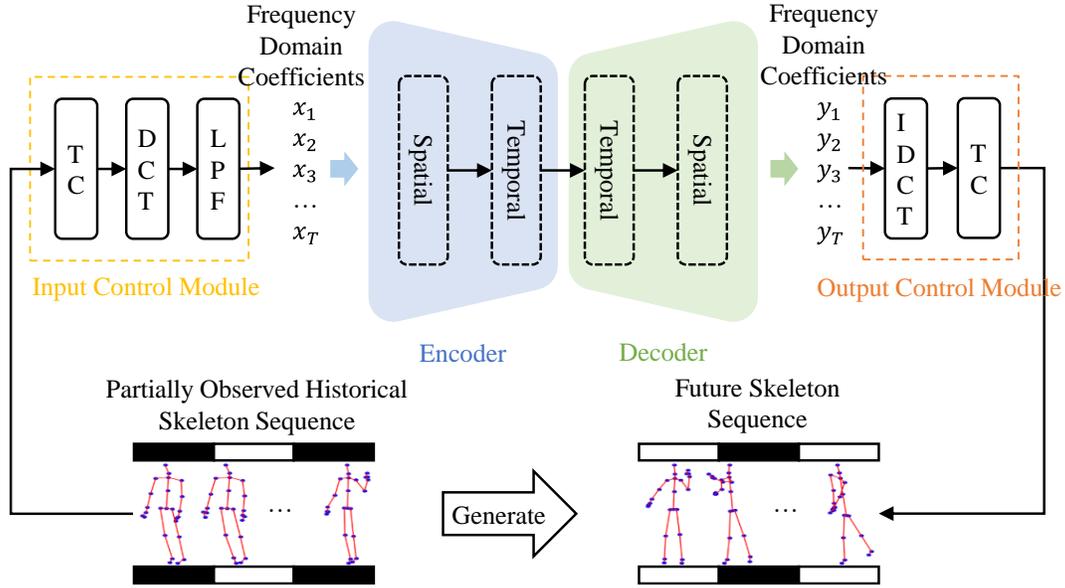

Figure 1 Overall Architecture Diagram of the Algorithm

## 2 Input/Output Control Module Based on Length Consistency

### 2.1 Input Control Module

The input control module consists of three parts: time control, discrete cosine transform, and low-pass filtering.

Firstly, the time control can cut or pad the historical skeleton sequence according to a set threshold, ensuring that the sequence input into the encoder has a consistent length. Suppose a human skeleton sequence with continuous T frames is represented as $X_{1:T} \in \mathbb{R}^{T \times K}$, with the duration threshold set to $T_{in}$. The defined time control is a binary function $C_{in}$: if $T \geq T_{in}$, only the last $T_{in}$ rows of the sequence are retained, i.e., the output is $X_{T-T_{in}+1:T}$; ; otherwise, stationary padding is applied to the sequence to supplement the data to $T_{in}$ rows.

The discrete cosine transform (DCT) can convert time-domain or spatial-domain signals to the frequency domain, thus having good decorrelation performance, and is commonly used for signal or image compression. Therefore, this paper also uses DCT to process historical skeleton sequences. Taking the motion trajectory of the action "throwing" (Figure 2) as an example, the green dashed line in the figure represents the motion trajectory of the left wrist joint, and the red dashed line represents the motion trajectory of the right wrist joint.

This section extracts the Y-direction spatial coordinates of the left wrist joint and right wrist joint, respectively, and plots the curves of the motion trajectories changing over time as shown in Figures 3(a) and 3(b). The frequency domain coefficients after DCT processing of the motion trajectories are shown in Figures 3(c) and 3(d). Although it can be seen from Figure 3 that the Y-direction motion amplitude of the right wrist joint is significantly greater than that of the left wrist joint, in the time-domain graph, it is necessary to observe the data of all frames to determine the difference between the maximum and minimum values. In the frequency-domain graph, only the coefficients of the low-frequency part need to be observed. Therefore, using DCT to process historical skeleton sequences can gather important information of the motion trajectory in one place (mid-low frequency part), while unimportant (high-frequency) frequency-domain coefficients can be directly discarded, achieving the purpose of dimension reduction and efficiency improvement.

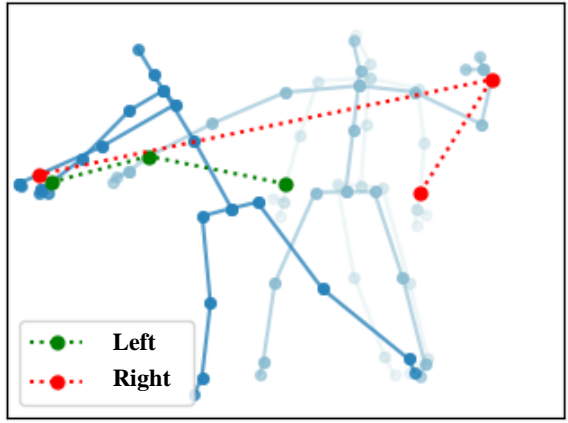

Figure 2 Motion Trajectory Diagram of the Activity 'Throwing'

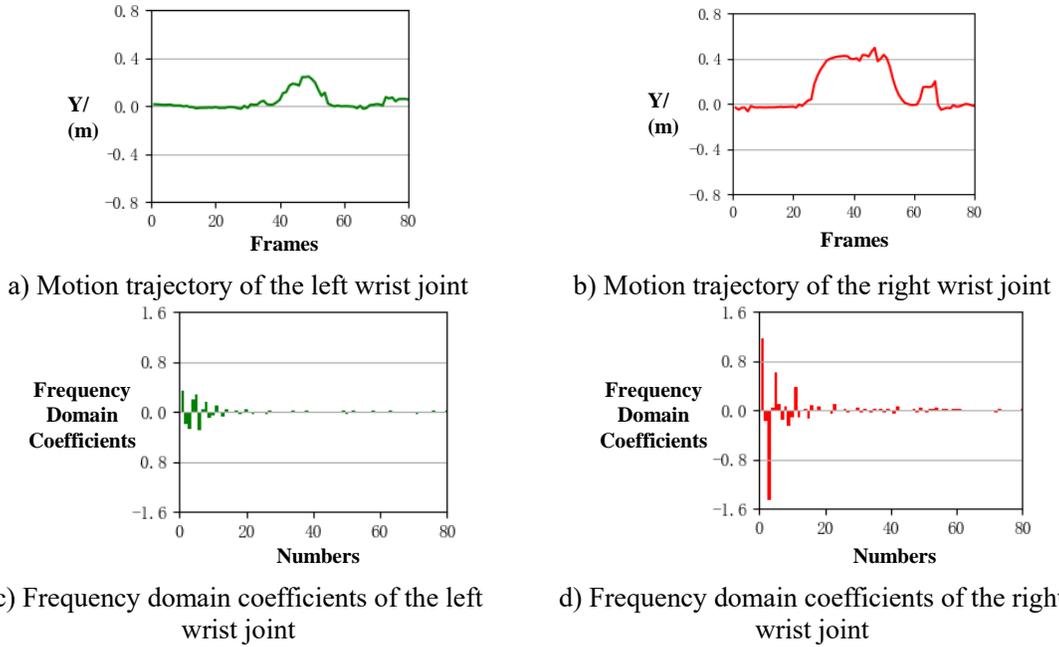

a) Motion trajectory of the left wrist joint

b) Motion trajectory of the right wrist joint

c) Frequency domain coefficients of the left wrist joint

d) Frequency domain coefficients of the right wrist joint

Figure 3 Time Variation Curves of Spatial Coordinates/Frequency Domain Coefficients of the Left and Right Wrist Joints

Low-Pass Filter (LPF) is defined as setting a cutoff frequency to block signals above that frequency, assigning them a value of zero. As shown in Figure 4, the majority of motion information is concentrated in the low-frequency part. Combined with the previous qualitative analysis, using a low-pass filter can effectively retain most important information and reduce the floating-point computational load of the network. The low-pass filter designed in this section is a binary function $\mathcal{F}_{LPF}$, mathematically expressed as:

$$\mathcal{F}_{LPF}(\lambda, \boldsymbol{C}_{1:T}) = \begin{bmatrix} \boldsymbol{C}_{1:\lfloor \lambda T \rfloor} \\ \boldsymbol{O} \end{bmatrix} \in \mathbb{R}^{T \times K} \tag{1}$$

where $\boldsymbol{C}_{1:T} \in \mathbb{R}^{T \times K}$ represents the frequency domain coefficient matrix, $\lambda (0 < \lambda \leq 1)$ represents the relative cutoff frequency. When $\lambda = 1$, it means all frequency domain coefficients are retained, $\lfloor \cdot \rfloor$ denotes the floor operation, $\boldsymbol{C}_{1:\lfloor \lambda T \rfloor} \in \mathbb{R}^{\lfloor \lambda T \rfloor \times K}$ represents the submatrix taking the first row to the $\lfloor \lambda T \rfloor$-th row of $\boldsymbol{C}_{1:T}$, and $\boldsymbol{O}$ represents the zero matrix.

## 2.2 Output Control Module

The output control module consists of two parts: inverse discrete cosine transform and time control.

Since discrete cosine transform (DCT) and inverse discrete cosine transform (IDCT) are lossless and symmetric transformations, they will not be specifically introduced. As the encoder-decoder network does not change the sequence's time length, the sequence length after IDCT will be the input duration threshold $T_{in}$. If the output duration threshold is set to $T_{out}$, then the time control in the output control module is a binary function $\mathcal{C}_{out}$: if $T_{in} \geq T_{out}$, only the first $T_{out}$ rows of the future sequence are retained; otherwise, all rows of the future sequence are retained, and then the generated future sequence is concatenated with the historical sequence and re-entered into the network. This process is repeated until the length of the retained data exceeds $T_{out}$, and finally, the first $T_{out}$ rows of data are taken as the output.

## 3 Encoder-Decoder Network Structure Based on Improved MLP

The encoder and decoder form a symmetrical structure, with the smallest unit being an improved MLP[17]. Compared to traditional MLP, the improved MLP replaces the activation function layer with an LN layer and adds residual connections, making it more suitable for Seq2Seq tasks.

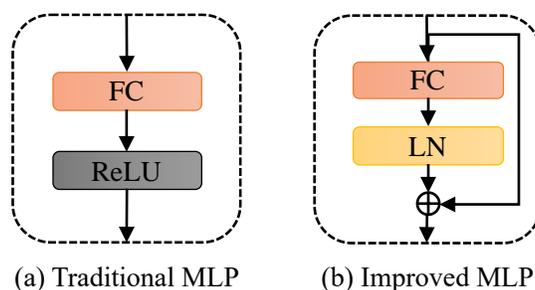

(a) Traditional MLP　　(b) Improved MLP
Figure 4 Differences between Traditional MLP and Improved MLP

Assuming the input duration control threshold is $T_{in}$, the partially observed historical skeleton sequence is represented as $\boldsymbol{X}_{1:T_{in}} \in \mathbb{R}^{T_{in} \times K}$. Assuming the output duration control threshold is $T_{out}$, the future skeleton sequence generated by the network is represented as $\boldsymbol{Y} \in \mathbb{R}^{T_{out} \times K}$.

### 3.1 Encoder

The input to the encoder is the frequency domain coefficient matrix $\boldsymbol{C}_{1:T_{in}} \in \mathbb{R}^{T_{in} \times K}$ processed by the input control module, i.e.,

$$\boldsymbol{E}^{(0)} = \boldsymbol{C}_{1:T_{in}} = \mathcal{F}_{LPF}\left(\lambda, \mathcal{D}(\boldsymbol{X}_{1:T_{in}})\right) \tag{2}$$

where $\boldsymbol{E}^{(0)}$ represents the input features of the encoder/spatial encoding. The encoder first models the spatial information of the skeleton sequence by concatenating $N_S$ improved MLPs to extract spatial features. The input/output channel number of the improved MLP is $K$. The iterative calculation formula for Spatial Encoding (SE) is

$$\boldsymbol{E}^{(i)} = \text{LN}_{SE}^{(i)}\left(\text{FC}_{SE}^{(i)}(\boldsymbol{E}^{(i-1)})\right) + \boldsymbol{E}^{(i-1)} \tag{3}$$

Besides spatial features, skeleton sequences also contain a large amount of temporal motion information. Therefore, after completing spatial information encoding, the encoder extracts temporal features by concatenating $N_T$ improved MLPs. The input/output channel number of the improved MLP is $T_{in}$. To facilitate temporal information modeling, a transpose operation is needed to convert the channels from the pose dimension to the time dimension, i.e.,

$$\widehat{\boldsymbol{E}}^{(0)} = \left(\boldsymbol{E}^{(N_S)}\right)^T \tag{4}$$

Where $\widehat{\boldsymbol{E}}^{(0)} \in \mathbb{R}^{K \times T_{in}}$ represents the input features of temporal encoding. The iterative calculation formula for Temporal Encoding (TE) is

$$\widehat{\boldsymbol{E}}^{(j)} = \text{LN}_{TE}^{(j)}\left(\text{FC}_{TE}^{(j)}(\widehat{\boldsymbol{E}}^{(j-1)})\right) + \widehat{\boldsymbol{E}}^{(j-1)} \tag{5}$$

### 3.2 Decoder

The structure of the decoder is completely symmetrical to that of the encoder. First, $N_T$ improved MLPs are concatenated to decode temporal features, and then $N_S$ improved MLPs decode spatial features to obtain the frequency domain coefficient matrix of the future skeleton sequence. Finally, the future skeleton sequence is obtained through the output control module. The input features of the decoder/temporal decoding are designed as

$$\widehat{D}^{(0)} = \widehat{E}^{(N_T)} \qquad (6)$$

The iterative calculation formula for Temporal Decoding (TD) is designed as

$$\widehat{D}^{(j)} = \text{LN}_{TD}^{(j)}\left(\text{FC}_{TD}^{(j)}(\widehat{D}^{(j-1)})\right) + \widehat{D}^{(j-1)} \qquad (7)$$

The input features of spatial decoding are designed as

$$D^{(0)} = \left(\widehat{D}^{(N_T)}\right)^T \qquad (8)$$

The iterative calculation formula for Spatial Decoding (SD) is designed as

$$D^{(i)} = \text{LN}_{SD}^{(i)}\left(\text{FC}_{SD}^{(i)}(D^{(i-1)})\right) + D^{(i-1)} \qquad (9)$$

The output of the decoder, processed by the output control module, produces the future skeleton sequence, calculated as

$$Y = C_{out}\left(T_{out}, \mathcal{D}^{-1}(D^{(N_S)})\right) \qquad (10)$$

In formulas (2) to (10), $1 \leq j \leq N_T$, $1 \leq i \leq N_S$.

## 4 Loss Function Design Based on Multi-Order Kinematic Features

The loss function $\mathcal{L}_{FSGN}$ of the future skeleton generation network proposed in this chapter consists of three parts: the mean per joint position error $\mathcal{L}_P$, the mean per joint velocity error $\mathcal{L}_V$, and the mean per joint acceleration error $\mathcal{L}_A$. The calculation formula is

$$\mathcal{L}_{FSGN} = \mathcal{L}_P + \alpha_V \mathcal{L}_V + \alpha_A \mathcal{L}_A \qquad (11)$$

where $\alpha_V$ and $\alpha_A$ are the velocity loss coefficient and acceleration loss coefficient (hyperparameters), respectively.

The mean per joint position error (MPJPE) is a core metric for evaluating the future skeleton generation network. The calculation formula is designed as

$$MPJPE = \frac{1}{V \times T}\sum_{v=1}^{V}\sum_{t=1}^{T}\|y_{t,v} - x_{t,v}\|_2 \qquad (12)$$

where $V$ represents the number of skeleton joints, $T$ represents the number of frames generated by the network, $y_{t,v} \in \mathbb{R}^3$ and $x_{t,v} \in \mathbb{R}^3$ represent the coordinates (velocity/acceleration) of the $v$-th joint in the $t$-th frame generated by the network and the actual future skeleton, respectively, and $\|\cdot\|_2$ denotes the 2-norm of a vector. Additionally, instead of constraining the network to directly generate future skeleton joint coordinates, this chapter constrains the network to generate the relative displacement of joint motion with respect to the last frame of the observed sequence. The future skeleton sequence coordinates are obtained indirectly by adding the displacements. Therefore, the calculation formula for $\mathcal{L}_p$ is designed as

$$\mathcal{L}_p = \frac{1}{V \times T}\sum_{v=1}^{V}\sum_{t=1}^{T}\|y_{t,v} - (z_v + x'_{t,v})\|_2 \qquad (13)$$

Where $z_v \in \mathbb{R}^3$ represents the coordinates of the $v$-th joint in the last frame of the observed sequence, and $x'_{t,v} \in \mathbb{R}^3$ represents the relative displacement of the $v$-th joint in the $t$-th frame generated by the network.

From a kinematic perspective, besides constraining joint motion displacement, joint motion velocity and acceleration are also important potential information. This chapter calculates joint motion velocity and acceleration using frame-by-frame differences. The calculation formula is designed as

$$\dot{X}_{1:T}[t,:] = X_{1:T}[t,:] - X_{1:T}[t-1,:] \qquad (14)$$

where $2 \leq t \leq T$, $X_{1:T}[t,:] \in \mathbb{R}^K$ represents the joint position (velocity) in the $t$-th frame, $\dot{X}_{1:T}[t,:] \in \mathbb{R}^K$ represents the joint velocity (acceleration) in the $t$-th frame, and $\dot{X}_{1:T}[1,:] = \mathbf{0} \in \mathbb{R}^K$. The calculation formulas for $\mathcal{L}_v$ and $\mathcal{L}_a$ are given in Equation (12).

# 5  Experiment

This section analyzes and verifies the effectiveness of the proposed methods for future synthesis and early prediction of production activities using publicly available datasets and datasets collected on-site in production environments.

## 5.1  Experimental Conditions

All experiments in this chapter were conducted on an RTX 2080 Ti graphics card. The deep neural network model proposed in this chapter was built using the PyTorch deep learning framework, and the experimental programs were run with the Python 3.7 interpreter on the Windows 10 operating system.

## 5.2  Datasets

(1) NJUST-3D
The NJUST-3D dataset[18, 19] simulates a manufacturing workshop environment in a laboratory and collects 18 common production activities. These activities are divided into four categories: early preparation activities (e.g., putting on work clothes, wearing a hard hat, releasing static electricity), production activities (e.g., moving materials, sawing workpieces, picking up parts), violation activities (e.g., violent handling, smoking, heavy hammering), and other activities (e.g., drinking water, wiping sweat, sitting). The NJUST-3D dataset uses two Microsoft Kinect V2 cameras to film 20 performers from different angles, collecting a total of 2160 samples (120 samples for each activity). Each sample contains four data modalities: RGB video, depth video, infrared image, and skeleton sequence (3D spatial coordinates of 25 skeleton joints).

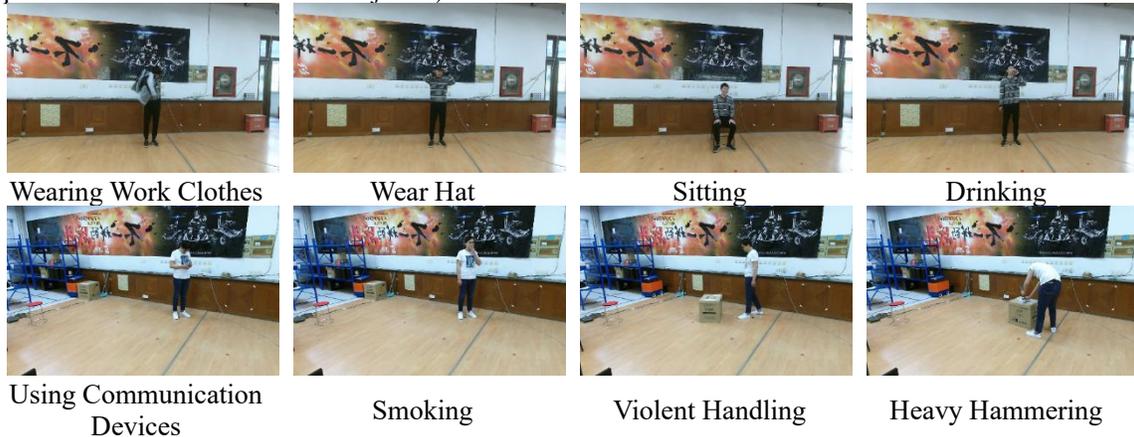

Wearing Work Clothes    Wear Hat    Sitting    Drinking

Using Communication Devices    Smoking    Violent Handling    Heavy Hammering

Figure 5 Some Activity Samples from the NJUST-3D Dataset

(2) Human3.6M
The Human3.6M dataset[20] is a large publicly available dataset for 3D human pose estimation research. It records approximately 3.6 million frames of human pose data and RGB images from a high-speed motion capture system, using four high-resolution cameras filming at 50 Hz. These poses come from 11 performers in 15 activities across 7 different scenarios, including directing, discussing, eating, greeting, calling, posing, shopping, sitting, smoking, taking photos, waiting, walking, dog walking, and walking together. Each scenario includes many types of variations, such as walking with hands in pockets, walking with a backpack, and walking while holding someone's hand. The human pose is composed of 32 joints.

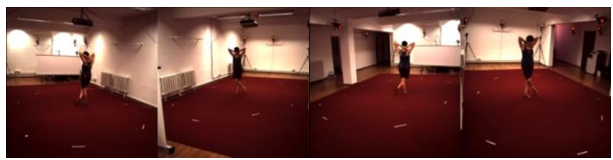

Waiting

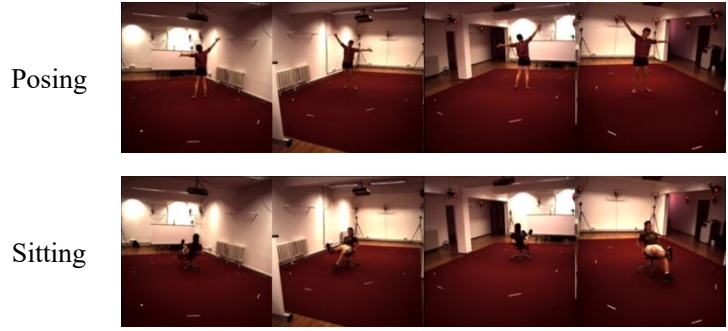

Posing

Sitting

Figure 6 Some Activity Samples from the Human3.6M Dataset

## 5.3 Effectiveness Verification Experiments on the Human3.6M Dataset

(1) Comparison with Other Advanced Algorithms

According to the evaluation protocol proposed in the literature[15], the Human3.6M dataset is downsampled to 25 FPS at equal intervals, using S1, S6, S7, S8, and S9 as the training set and S5 as the test set. During the testing phase, 256 random samples of each activity from the test set are selected to evaluate the MPJPE of 22 joints within 1000 ms (25 frames). The hyperparameter settings of the future skeleton generation network are shown in Table 1, and all hyperparameters are determined by ablation experiments. The training loss curve of the future skeleton generation network is shown in Figure 7, with a total of 80 training rounds and approximately 55,000 parameter updates through backpropagation.

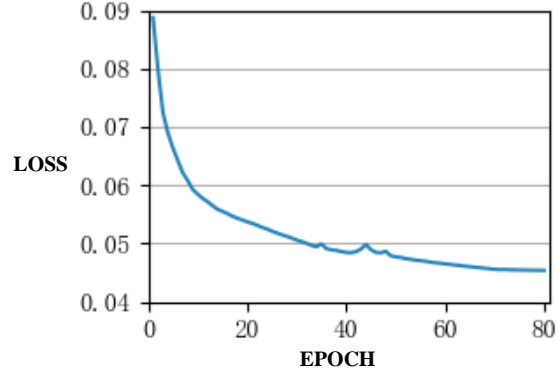

Figure 7 Training Loss Curve of the Future Skeleton Generation Network on the Human3.6M Dataset

Table 1 Hyperparameter Settings of the Future Skeleton Generation Network

| Variable | Definitions | Value |
| --- | --- | --- |
| $T_{in}$ | Input Control Threshold | 50 |
| $T_{out}$ | Output Control Threshold | 10 |
| $\lambda$ | Relative Cutoff Frequency | 0.8 |
| $N_S$ | Amount of Spatial Encoding/Decoding Units | 1 |
| $N_T$ | Amount of Temporal Encoding/Decoding Units | 20 |
| $\alpha_V$ | Velocity Loss Coefficient | 1.0 |
| $\alpha_A$ | Acceleration Loss Coefficient | 1.0 |

This chapter reports the MPJPE in millimeters for the future skeleton generation network at different time steps and compares it with the reported values of other advanced algorithms, as shown in Table 2. The MPJPE at each time step is better than existing methods, including RNN-based methods[12], CNN-based methods[13], and GCN-based methods[14-16], and the model has fewer parameters compared to existing methods. Figure 8 shows the MPJPE at each time step. The experimental results indicate that the network matches real motion conditions in short-term generation (0-400 ms) and roughly conforms to real motion conditions in long-term generation (400-1000 ms). However, when generating for longer durations, the error increases rapidly because the network uses iterative regression to generate a longer future, leading to error accumulation, and the uncertainty in generating future motion trajectories grows quickly over time."

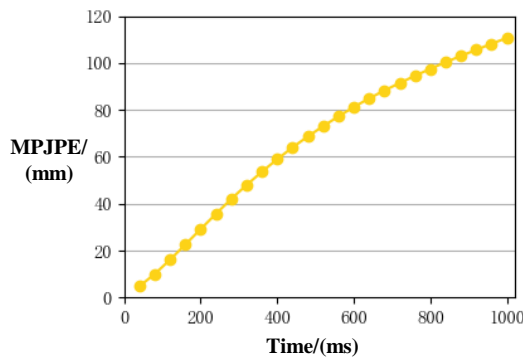

Figure 8 Experimental Results of the Future Skeleton Generation Network on the Human3.6M Dataset

Table 2 Experimental Results of the Future Skeleton Generation Network on the Human3.6M Dataset

| Methods | Parameters/M | MPJPE@Time/mm | | | | | | | |
|---|---|---|---|---|---|---|---|---|---|
| | | 80 | 160 | 320 | 400 | 560 | 720 | 880 | 1000 |
| Res-RNN[12] | 3.44 | 25.0 | 46.2 | 77.0 | 88.3 | 106.3 | 119.4 | 130.0 | 136.6 |
| convSeq2Seq[13] | 15.58 | 16.6 | 33.3 | 61.4 | 72.7 | 90.7 | 104.7 | 116.7 | 124.2 |
| LTD-GCN[14] | 2.55 | 11.2 | 23.4 | 47.9 | 58.9 | 78.3 | 93.3 | 106.0 | 114.0 |
| Hisrep[15] | 3.24 | 10.4 | 22.6 | 47.1 | 58.3 | 77.3 | 91.8 | 104.1 | 112.1 |
| MSR-GCN[16] | 6.30 | 11.3 | 24.3 | 50.8 | 61.9 | 80.0 | —— | —— | 112.9 |
| FSGN | **0.12** | **10.1** | **22.6** | **48.0** | **59.0** | **77.2** | **91.4** | **103.0** | **110.6** |

(2) Ablation Experiments

This chapter evaluates the impact of different factors on the future skeleton generation network using the Human3.6M dataset:

**a. Input/Output Control Thresholds**

As shown in Table 3, this chapter ablated the input/output control thresholds. The experimental results indicate that a larger observation amount $T_{in}$ is beneficial for long-term generation but increases the short-term generation error because short-term generation relies more on the most recently observed data. Increasing $T_{in}$ reduces the network's focus on the latest data. Generating future skeleton sequences frame by frame ($T_{out} = 1$) results in a significant error accumulation effect, whereas one-shot mapping of future skeleton sequences ($T_{out} = 25$) performs worse than multiple iterative generations. Therefore, the hyperparameters are set to $T_{in} = 50$, $T_{out} = 10$.

Table 3 Ablation Results of Input/Output Control Thresholds

| $T_{in}$ | $T_{out}$ | MPJPE@Time/ms | | | | | | | |
|---|---|---|---|---|---|---|---|---|---|
| | | 80 | 160 | 320 | 400 | 560 | 720 | 880 | 1000 |
| 25 | 10 | 10.7 | 23.9 | 50.8 | 62.3 | 81.3 | 96.2 | 107.5 | 114.3 |
| 50 | 10 | **10.1** | **22.6** | **48.0** | **59.0** | **77.2** | **91.4** | 103.0 | 110.6 |
| 75 | 10 | 10.4 | 23.0 | 48.8 | 59.9 | 77.8 | 91.5 | **102.3** | **109.7** |
| 50 | 1 | 10.4 | 23.4 | 49.9 | 61.6 | 80.6 | 95.7 | 108.4 | 116.9 |
| 50 | 25 | 12.9 | 25.8 | 52.0 | 63.0 | 80.6 | 94.0 | 104.9 | 112.0 |

**b. Amounts of Encoding/Decoding Units**

As shown in Table 4, this chapter ablated the amounts of temporal/spatial encoding/decoding units. The experimental results indicate that increasing the number of spatial encoding/decoding units $N_S$ degrades the skeleton generation performance. The number of temporal encoding/decoding units $N_T$ achieves the best performance at 16 and 20, with $N_T = 20$ yielding lower long-term generation errors. Therefore, the hyperparameters are set to $N_S = 1$, $N_T = 20$.

Table 4 Ablation Results of the Number of Encoding/Decoding Units

| $N_S$ | $N_T$ | MPJPE@Time/ms | | | | | | | |
|---|---|---|---|---|---|---|---|---|---|
| | | 80 | 160 | 320 | 400 | 560 | 720 | 880 | 1000 |
| 1 | 1 | 12.1 | 27.0 | 56.6 | 68.8 | 88.1 | 101.5 | 112.1 | 119.1 |
| 1 | 4 | 11.1 | 24.5 | 51.8 | 63.4 | 81.8 | 95.8 | 107.4 | 115.1 |
| 1 | 16 | **10.0** | **22.5** | **48.0** | 59.1 | 77.4 | 91.7 | 103.5 | 111.4 |
| 1 | 20 | 10.1 | 22.6 | **48.0** | **59.0** | **77.2** | **91.4** | **103.0** | **110.6** |
| 1 | 24 | 10.2 | 22.9 | 48.6 | 59.6 | 77.6 | 91.6 | 103.3 | 110.9 |
| 1 | 28 | 10.4 | 23.2 | 49.2 | 60.4 | 78.8 | 93.2 | 104.8 | 112.3 |
| 1 | 32 | 10.5 | 23.2 | 48.9 | 59.8 | 77.9 | 92.4 | 104.1 | 111.7 |
| 2 | 20 | 11.5 | 24.3 | 50.0 | 61.1 | 79.2 | 93.1 | 104.8 | 112.4 |
| 4 | 20 | 23.8 | 44.4 | 76.8 | 89.3 | 108.1 | 122.1 | 132.1 | 137.3 |

### c. (Acceleration) Loss Coefficient of the Loss Function

As shown in Table 5, this chapter explores the impact of the (acceleration) loss coefficient. The experimental results indicate that the network can converge well without adding additional loss function terms, constraining only the mean per joint position error (relative displacement). When the velocity loss coefficient $\alpha_V = 1.0$ and the acceleration loss coefficient $\alpha_A = 1.0$, the network achieves the best convergence.

Table 5 Ablation Results of the (Acceleration) Loss Coefficient

| $\alpha_V$ | $\alpha_A$ | MPJPE@Time/mm | | | | | | | |
|---|---|---|---|---|---|---|---|---|---|
| | | 80 | 160 | 320 | 400 | 560 | 720 | 880 | 1000 |
| 0 | 0 | **10.1** | **22.6** | **48.0** | 59.1 | 77.9 | 92.9 | 104.5 | 112.0 |
| 0.5 | 0 | **10.1** | **22.6** | 48.1 | **59.0** | 77.3 | 91.7 | 103.7 | 111.8 |
| 1.0 | 0 | **10.1** | **22.6** | 48.2 | 59.1 | **77.2** | 91.5 | 103.1 | 111.1 |
| 2.0 | 0 | 10.8 | 23.8 | 49.3 | 60.1 | 78.0 | 92.3 | 104.1 | 112.1 |
| 1.0 | 0.5 | **10.1** | 22.8 | 48.5 | 59.4 | 77.5 | 91.5 | 103.1 | 110.7 |
| 1.0 | 1.0 | **10.1** | **22.6** | **48.0** | **59.0** | **77.2** | **91.4** | **103.0** | **110.6** |
| 1.0 | 2.0 | 10.2 | 22.9 | 48.7 | 59.9 | 78.1 | 92.4 | 104.1 | 111.7 |

### d. Network Components

As shown in Table 6, this chapter ablated various components of the future skeleton generation network, including spatial units, temporal units, layer normalization of the improved MLP, and DCT and its inverse transform. The experimental results demonstrate that all network components are indispensable, with the most significant impact from DCT and its inverse transform, and the least impact from the layer normalization of the improved MLP. Even when the encoder-decoder is purely linear transformation, modeling motion trajectories through DCT achieves good generation results, such as the MPJPE@1000ms without layer normalization increasing by only 17.6mm. To validate the effectiveness of the improved MLP proposed in the literature[17], this chapter also compared the generation effect using traditional MLP. The results show that traditional ReLU-based MLP in the encoder-decoder framework cannot effectively model the spatiotemporal patterns of motion trajectories.

Table 6 Ablation Results of Network Components

| Ablation | MPJPE@Time/mm | | | | | | | |
|---|---|---|---|---|---|---|---|---|
| | 80 | 160 | 320 | 400 | 560 | 720 | 880 | 1000 |
| No Spatial Unit | 21.2 | 34.5 | 63.2 | 75.1 | 98.8 | 110.7 | 127.6 | 133.6 |
| No Temporal Unit | 23.8 | 43.8 | 75.9 | 88.4 | 106.3 | 120.4 | 130.0 | 135.8 |
| No LN | 13.1 | 29.1 | 62.5 | 76.7 | 97.9 | 111.9 | 122.1 | 128.2 |
| No DCT | 27.8 | 48.1 | 79.9 | 92.2 | 113.3 | 127.5 | 138.9 | 145.4 |
| Traditional MLP | 23.8 | 44.4 | 76.8 | 89.3 | 108.1 | 122.1 | 132.1 | 137.3 |
| FSGN | **10.1** | **22.6** | **48.0** | **59.0** | **77.2** | **91.4** | **103.0** | **110.6** |

### e. Relative Cutoff Frequency of Low-Pass Filtering

As shown in Table 7, this chapter ablated the relative cutoff frequency of low-pass filtering. The experimental results indicate that when the relative cutoff frequency λ = 0.8, the MPJPE within 720ms is better than the results for λ = 1.0. Therefore, appropriately discarding some high-frequency information can remove noise introduced during data collection, which helps the model learn the spatiotemporal dynamics of motion trajectories. As the relative cutoff frequency continues to decrease, the generation effect gradually worsens due to the loss of key motion information. When λ = 0.2, the MPJPE at 1000ms increases by 8.3mm. The results demonstrate that DCT can efficiently model temporal information; even when discarding 80% of the data, the generation error is still smaller than without using DCT.

Table 7 Ablation Results of Relative Cutoff Frequency

| λ | MPJPE@Time/mm | | | | | | | |
|---|---|---|---|---|---|---|---|---|
| | 80 | 160 | 320 | 400 | 560 | 720 | 880 | 1000 |
| 1.0 | 10.4 | 23.6 | 50.4 | 61.7 | 79.6 | 92.3 | **102.8** | **109.9** |
| 0.8 | **10.1** | **22.6** | **48.0** | **59.0** | **77.2** | **91.4** | 103.0 | 110.6 |
| 0.6 | 11.6 | 25.3 | 51.9 | 63.0 | 80.4 | 93.2 | 103.6 | 110.8 |
| 0.4 | 14.7 | 29.0 | 54.9 | 65.5 | 82.3 | 94.6 | 105.2 | 112.4 |
| 0.2 | 26.7 | 42.0 | 66.8 | 76.5 | 90.9 | 102.0 | 112.3 | 118.5 |

**f. Relative Displacement of the Loss Function**

As shown in Table 8, this chapter ablated the impact of relative displacement in the loss function. The experimental results indicate that directly generating the absolute position coordinates of skeleton joints results in larger errors. The MPJPE@1000ms of absolute positions increases by 13.1mm compared to relative displacements, proving that indirectly generating future skeleton sequences is an effective strategy as proposed in the literature[17].

Table 8 Ablation Results of Relative Displacement

| Ablation | MPJPE@Time/mm | | | | | | | |
|---|---|---|---|---|---|---|---|---|
| | 80 | 160 | 320 | 400 | 560 | 720 | 880 | 1000 |
| Absolute Position | 12.7 | 28.5 | 59.7 | 72.1 | 93.6 | 107.0 | 116.8 | 123.7 |
| Relative Displacement | **10.1** | **22.6** | **48.0** | **59.0** | **77.2** | **91.4** | **103.0** | **110.6** |

## 5.3 Effectiveness Verification Experiments on the NJUST-3D Dataset

The NJUST-3D dataset is a small-scale dataset with short sequences. Figure 9 shows the histogram of the sample time lengths in the dataset, with an average time length of 95.9 frames and a total of 2160 original samples. Analogous to the evaluation protocol proposed in the literature[15], this paper customizes the evaluation protocol for the NJUST-3D dataset as follows: samples labeled R002 and R003 are used as the training set, and samples labeled R001 are used as the test set. During the testing phase, 256 random samples of each activity from the test set are selected to evaluate the MPJPE of 25 joints within 500 ms (15 frames). Based on empirical knowledge, the hyperparameter settings for the future skeleton generation network trained on the NJUST-3D dataset are shown in Table 1, with modifications to $T_{in} = 30$ and $T_{out} = 6$. The training loss curve is shown in Figure 10, with a total of 80 training rounds and approximately 28,000 parameter updates through backpropagation.

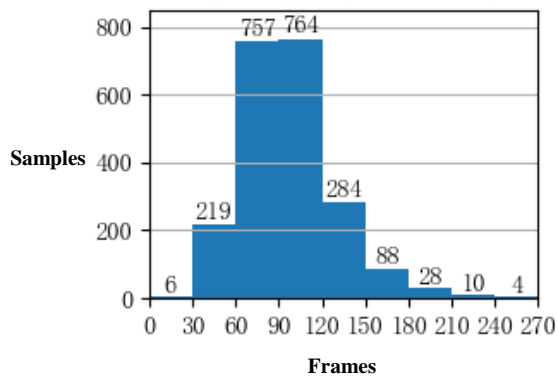

Figure 9 Histogram of Sample Time Lengths in the NJUST-3D Dataset

This section reports the MPJPE in millimeters for the future skeleton generation network at different time steps, as shown in Table 9. It should be noted that due to occlusion and other reasons, the skeleton data for the activities "using a cart to transport" and "violent material handling" have significant missing and drifting data, leading to noticeable random fluctuations in the network's generation error. Therefore, when calculating the MPJPE for the test set, these two abnormal data groups were excluded, and the average MPJPE of the remaining 16 activities was reported.

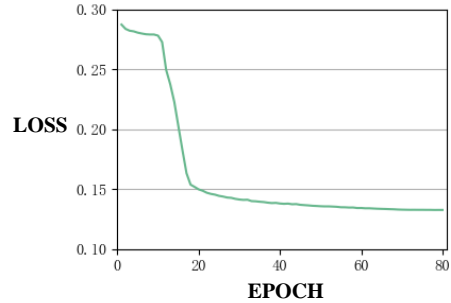

Figure 10 Training Loss Curve of the Future Skeleton Generation Network on the NJUST-3D Dataset

Table 9 Experimental Results of the Future Skeleton Generation Network on the NJUST-3D Dataset

| | MPJPE@Time/mm | | | | |
|---|---|---|---|---|---|
| | 100 | 200 | 300 | 400 | 500 |
| FSGN | 48.3 | 64.7 | 83.5 | 92.2 | 106.1 |

## 5 Conclusions

This paper addresses the issue of existing future skeleton synthesis methods relying on deep learning models with high computational complexity by proposing a lightweight encoder-decoder framework for the future skeleton generation network. The network solves the problem of variable time lengths for input/output sequences using an input/output control module based on threshold control and (inverse) discrete cosine transform. The encoder-decoder framework, based on spatio-temporal encoding and decoding units, maps historical sequences to future sequences and trains the network by constraining the relative displacement, velocity, and acceleration of joint movements. Experiments on the Human3.6M dataset and the NJUST-3D dataset demonstrate that the proposed network has lower generation error and fewer model parameters compared to existing methods, providing supplementary future information for subsequent early activity prediction.

The experimental results indicate that the proposed future skeleton synthesis algorithm performs better on the Human3.6M and NJUST-3D datasets compared to existing methods, with lower generation errors and fewer model parameters, effectively supplementing future information for subsequent early activity prediction. Ablation experiments also verify the importance of each component in the network.

## Acknowledgments

This work was supported in part by The Key Project of Certain Research Program (No. JCKY2020210B006 and No. JCKY2017204B053), and The National Key Research and Development Program of China (No. 2020YFB1708400).

## References

[1] HU X J, DAI J Z, LI M, et al. Online human action detection and anticipation in videos: A survey [J]. Neurocomputing, 2022, 491: 395-413.
[2] KONG Y, KIT D, FU Y. A Discriminative Model with Multiple Temporal Scales for Action Prediction; proceedings of the 13th European Conference on Computer Vision (ECCV), Zurich, SWITZERLAND, F Sep 06-12, 2014 [C]. Springer International Publishing Ag: CHAM, 2014.
[3] KONG Y, FU Y. Max-margin action prediction machine [J]. IEEE transactions on pattern analysis

and machine intelligence, 2015, 38(9): 1844-58.
[4] WANG X H, HU J F, LAI J H, et al. Progressive Teacher-student Learning for Early Action Prediction; proceedings of the 32nd IEEE/CVF Conference on Computer Vision and Pattern Recognition (CVPR), Long Beach, CA, F Jun 16-20, 2019 [C]. Ieee Computer Soc: LOS ALAMITOS, 2019.
[5] LIU C W, ZHAO X X, LI Z K, et al. A Novel Two-Stage Knowledge Distillation Framework for Skeleton-Based Action Prediction [J]. IEEE Signal Processing Letters, 2022, 29: 1918-22.
[6] GOU J P, YU B S, MAYBANK S J, et al. Knowledge Distillation: A Survey [J]. International Journal of Computer Vision, 2021, 129(6): 1789-819.
[7] RYOO M S. Human activity prediction: Early recognition of ongoing activities from streaming videos[C]//2011 international conference on computer vision. IEEE, 2011: 1036-1043.
[8] LI Z, ZHANG H B, ZHANG M H, et al. Late feature supplement network for early action prediction [J]. Image and Vision Computing, 2022, 125: 8.
[9] Gammulle H, Denman S, Sridharan S, et al. Predicting the future: A jointly learnt model for action anticipation[C]//Proceedings of the IEEE/CVF International Conference on Computer Vision. 2019: 5562-5571.
[10] Chen L, Lu J, Song Z, et al. Recurrent semantic preserving generation for action prediction[J]. IEEE Transactions on Circuits and Systems for Video Technology, 2020, 31(1): 231-245.
[11] Sun Z, Ke Q, Rahmani H, et al. Human action recognition from various data modalities: A review[J]. IEEE transactions on pattern analysis and machine intelligence, 2022, 45(3): 3200-3225.
[12] Martinez J, Black M J, Romero J. On human motion prediction using recurrent neural networks[C]//Proceedings of the IEEE conference on computer vision and pattern recognition. 2017: 2891-2900.
[13] Li C, Zhang Z, Lee W S, et al. Convolutional sequence to sequence model for human dynamics[C]//Proceedings of the IEEE conference on computer vision and pattern recognition. 2018: 5226-5234.
[14] MAO W, LIU M M, SALZMANN M, et al. Learning Trajectory Dependencies for Human Motion Prediction; proceedings of the 17th IEEE/CVF International Conference on Computer Vision (ICCV), F 2019, 2019 [C].
[15] Mao W, Liu M, Salzmann M, et al. Learning trajectory dependencies for human motion prediction[C]//Proceedings of the IEEE/CVF international conference on computer vision. 2019: 9489-9497.
[16] Dang L, Nie Y, Long C, et al. Msr-gcn: Multi-scale residual graph convolution networks for human motion prediction[C]//Proceedings of the IEEE/CVF international conference on computer vision. 2021: 11467-11476.
[17] Guo W, Du Y, Shen X, et al. Back to mlp: A simple baseline for human motion prediction[C]//Proceedings of the IEEE/CVF winter conference on applications of computer vision. 2023: 4809-4819.
[18] LIU T, LU Z, SUN Y, et al. Working activity recognition approach based on 3D deep convolutional neural network [J]. Computer Integrated Manufacturing Systems, 2020, 26(8): 2143-56..
[19] LIU T, HONG Q, SUN Y, et al. Approach for recognizing production action in digital twin shop-floor based on graph convolution network [J]. Computer Integrated Manufacturing Systems, 2021, 27(2): 501-9.
[20] Ionescu C, Papava D, Olaru V, et al. Human3. 6m: Large scale datasets and predictive methods for 3d human sensing in natural environments[J]. IEEE transactions on pattern analysis and machine intelligence, 2013, 36(7): 1325-1339.